\documentclass[twocolumn, switch]{article} % Method A for two-column formatting

\usepackage{preprint}

\usepackage{amsmath, amsthm, amssymb, amsfonts}

\usepackage[numbers,square]{natbib}
\usepackage[utf8]{inputenc}	% allow utf-8 input
\usepackage[T1]{fontenc}	% use 8-bit T1 fonts
\usepackage{xcolor}		% colors for hyperlinks
\usepackage[colorlinks = true,
            linkcolor = purple,
            urlcolor  = blue,
            citecolor = cyan,
            anchorcolor = black]{hyperref}	% Color links to references, figures, etc.
\usepackage{booktabs} 		% professional-quality tables
\usepackage{nicefrac}		% compact symbols for 1/2, etc.
\usepackage{microtype}		% microtypography
\usepackage{lineno}		% Line numbers
\usepackage{float}			% Allows for figures within multicol

\usepackage{lipsum}		%  Filler text

\usepackage{newfloat}
\DeclareFloatingEnvironment[name={Supplementary Figure}]{suppfigure}
\usepackage{sidecap}
\sidecaptionvpos{figure}{c}

\usepackage{titlesec}
\titlespacing\section{0pt}{12pt plus 3pt minus 3pt}{1pt plus 1pt minus 1pt}
\titlespacing\subsection{0pt}{10pt plus 3pt minus 3pt}{1pt plus 1pt minus 1pt}
\titlespacing\subsubsection{0pt}{8pt plus 3pt minus 3pt}{1pt plus 1pt minus 1pt}

\usepackage{tikz,xcolor,hyperref}

\definecolor{lime}{HTML}{A6CE39}
\DeclareRobustCommand{\orcidicon}{
	\begin{tikzpicture}
	\draw[lime, fill=lime] (0,0) 
	circle [radius=0.16] 
	node[white] {{\fontfamily{qag}\selectfont \tiny ID}};
	\draw[white, fill=white] (-0.0625,0.095) 
	circle [radius=0.007];
	\end{tikzpicture}
	\hspace{-2mm}
}
\foreach \x in {A, ..., Z}{\expandafter\xdef\csname orcid\x\endcsname{\noexpand\href{https://orcid.org/\csname orcidauthor\x\endcsname}
			{\noexpand\orcidicon}}
}
\title{Multimodality in Online Education: A Comparative Study}

\usepackage{authblk}

\author[1]{Praneeta Immadisetty$^*$}
\author[1]{Pooja Rajesh$^*$}
\author[1]{Akshita Gupta$^*$}
\author[2]{Anala M R}
\author[1]{Soumya A}
\author[3]{K. N. Subramanya}

\affil[1]{Department of Computer Science, RV College of Engineering}
\affil[2]{Department of Information Science, RV College of Engineering}
\affil[3]{RV College of Engineering}

\begin{document}

\twocolumn[ % Method A for two-column formatting
    \begin{@twocolumnfalse} % Method A for two-column formatting
  
    \maketitle

    \begin{abstract}
    The commencement of the decade brought along with it a grave pandemic and in response the movement of education forums predominantly into the online world. With a surge in the usage of online video conferencing platforms and tools to better gauge student understanding, there needs to be a mechanism to assess whether instructors can grasp the extent to which students understand the subject and their response to the educational stimuli. The current systems consider only a single cue with a lack of focus in the educational domain. Thus, there is a necessity for the measurement of an all-encompassing holistic overview of the students' reaction to the subject matter. This paper highlights the need for a multimodal approach to affect recognition and its deployment in the online classroom while considering four cues, posture and gesture, facial, eye tracking and verbal recognition. It compares the various machine learning models available for each cue and provides the most suitable approach given the available dataset and parameters of classroom footage. A multimodal approach derived from weighted majority voting is proposed by combining the most fitting models from this analysis of individual cues based on accuracy, ease of procuring data corpus, sensitivity and any major drawbacks.
    \end{abstract}

    \vspace{0.35cm}

    \end{@twocolumnfalse} % Method A for two-column formatting
] % Method A for two-column formatting

{
  \renewcommand{\thefootnote}%
    {\fnsymbol{footnote}}
  \footnotetext[1]{These authors contributed equally to this work}
}
%\begin{multicols}{2} % Method B for two-column formatting (doesn't play well with line numbers), comment out if using method A

%%%%%%%%%%%%%%%  Main text   %%%%%%%%%%%%%%%
% \linenumbers

\section{Introduction}
Affect Recognition or Emotion Recognition refers to the ability to identify non verbal (NV) cues such as facial expressions, body language, tone of speech and physiological signal interpretation. Emotion recognition becomes an essential parallel form of communication apart from the most widely used method of verbal communication. It helps to identify the tone of communication and helps allot context to the individual's thought process. In certain situations where true affective status is required to be identified, non verbal behaviour is more accurate in comparison to verbal behaviour. In fact, research has found that tone of speech is the most difficult cue to consciously control, thereby reinforcing the need to assist verbal communication with NV cues for complete context identification and emotion recognition \cite{Falk17}. Verbal cues are vocalised opinions made consciously. Verbal cues pose a threat of biassed feedback, whereas extracting analytics from NV cues eliminates this issue. However, to identify emotions based on these NV cues is a complex task. NV cues refer to cues that are involuntary. They need to be analysed before reaching any deterministic value.

The aforementioned NV cues however, are not as accurate and could prove to be ambiguous when analysed individually. Since emotions can have a wide range, a single cue may not suffice for the most cohesive affect recognition. For this reason, there is a need to combine multiple modalities for better assessment of human emotions. Employing multiple cues for identifying the emotional state leads to a more robust and realistic output, which is closer to a real-world scenario. The parameters that can contribute to the culmination of NV cues include posture, gesture, eye-tracking, speech recognition, facial recognition. These parameters together comprise the different modalities of Multimodal Affect Recognition (MAR). Moreover, given their non deterministic and involuntary nature, these cues cannot be modified or hampered with, and must lead to an accurate and unbiased feedback. Seeing emotion recognition from a multimodal perspective ensures the lack of domination of a singular characteristic that promises a more sensitive reaction to human engagement. In some scenarios, where certain cues cannot be properly identified or used, it becomes essential to combine the outputs from the other cues in a deterministic manner to facilitate the complete understanding of the emotion.

There have been several attempts to use emotions as a medium of deeper understanding under various circumstances. This includes affect recognition in classroom environments, driving simulations, autonomous cars, analysing visual media appeal, medical diagnosis, etc. Thus, emotion recognition becomes vital in fields where one cue, in most cases, verbal communication, is not sufficient to gather information. These situations include students of younger age groups who haven’t fully developed means to communicate their needs, special needs patients and also online education. Emotions are a vital source for understanding the extent of students’ engagement in online education. They can be incremental in determining the response of students in an online system. The psychological evaluation and emotional response of students is easy to gauge in an offline environment. However the clarity of video footage and difficulty of monitoring numerous screens simultaneously pose to be the major setbacks of e-learning. This leads to the lack of genuine and unbiased feedback for the teachers which hampers the ability to adapt to student reactions in real time. If a system that uses in-system footage instead of that over the web could be devised, it will provide better educational outcomes in terms of feedback generated.

To corroborate the problem in online education, it can be illustrated with the following instance. When a student joins a video conferencing platform for their online class, to facilitate bandwidth issues or in sheer disinterest, the student is allowed to turn off their camera. Hence the only form of communication between the student and the teacher is elicited verbal confirmation from the teacher of understanding the concept, or ensuring compulsory turning on of cameras. This is very different from offline education, where the environment to learn is more conducive and free from distractions. In addition, the students body language, eye tracking can all be assessed by the teacher. But in online education, such focused observations are more difficult to formulate. MAR can be helpful in such a situation, where a single cue like eye-tracking may not be sufficient, as a student could be looking at the screen, but the posture of their body may suggest disinterest in the class content. Hence, combining multiple cues could prove to be a better solution and provide a way to affirm class engagement similar to how it would be in an offline classroom.

In a survey conducted amongst 29 professors and 89 students from various fields of education, it was found that approximately 96\% of the interviewed students and teachers felt that remote learning was not completely effective for them. Moreover, only 15\% of the student respondents said that they provide honest feedback when asked for, and 94\% of them felt the need for the generation of honest feedback systems. Furthermore, 92\% of them felt that a system in which teachers could update their teaching methodology in accordance with the classroom interest levels would greatly improve the overall classroom experience. 

\section{Related Works}
This paper considers 4 major measures. The cues are based on the available parameters from an online classroom environment, namely, audio and video footage. Based on these available measures, it is possible to extract facial, gesture, posture, eye tracking and speech or verbal recognition. In facial recognition, the cues of various emotions presented facially are identified. These emotions can be divided into 6 categories: confused, frustrated, distracted, bored, interested and neutral. Gesture recognition encompassed all hand movements and body signs that can be recorded on video. This encompasses gestures such as hand over face, writing, nodding and tilting the head. Posture describes the overall body language of the student and can be understood by analysing slumped back, back straight, tilted away from the screen. Eye Tracking recognizes looking away, paying attention and zoning out (No pupil movement) as parameters. Using audio detecting yawning, mumbling, whining, and noise from other tabs is feasible. This section addresses the state of the art in these domains.

\subsection{GESTURE AND POSTURE RECOGNITION}
One of the earliest approaches for gesture classification was the Hidden Markov Model (HMM) \cite{10.1007/3-540-46616-9_10}. The HMM model is a statistical model comprising two states: hidden states and observable states. The model has an influx of a sequence of pre-defined gestures and based on the order, predictions of the incoming gestures are made. This was followed by a rise in Support Vector Machine (SVM) approaches \cite{5403381}. SVM can be implemented as a multiclass classifier by breaking it down into several binary classification problems. With the development of Kinect, there was an observable increase in the accuracy of recognising gestures. Kinect can be used for gauging the distance of an object from a Kinect camera by using 3D motion sensing. Facial features and gaze can be detected using the Kinect One sensor. \cite{Zaletelj} uses the Kinect One sensor to estimate student attention levels in the classroom by utilising features that were computed using the data procured. With the surge in popularity of the Convolutional Neural Network (CNN) around the early 2010s, there is a rise observed in its use to classify images or a series of images. \cite{8821186} uses a CNN model that employs the automatic extraction of class specific features and reduces the necessity of extracting them manually. It requires lesser pre-processing, lower training time, and a smaller dataset, which makes it a suitable solution for most classification problems. Head-over-face gestures as highlighted in \cite{Behera2020} can be extremely useful in determining the engagement levels of students and assessing the quality of subject matter being taught without the use of verbal cues. Table \ref{lit-gp} has a brief overview of the various approaches taken for emotion recognition using facial expressions.

\begin{table}
\caption{\textbf{Related Works for Gesture and Posture}}
\label{lit-gp}
\tiny
\setlength\tabcolsep{1.5pt}
\begin{tabular}{ |p{1cm}|p{1.25cm}|p{1.25cm}|p{2.25cm}|p{2.25cm}| }
\hline
\textbf{Author} & \textbf{Tools/Models} & \textbf{Dataset} & \textbf{Merits} & \textbf{Demerits} \\
\hline
Wu \textit{et al. 1999} \cite{10.1007/3-540-46616-9_10} & Elastic graph matching, DNF, HMM & N/A & Review on static and temporal hand gestures and techniques used for classifying them. & Focuses primarily on Hand Gestures, inadequate for understanding emotional state in educational context. \\
\hline
Wilson \textit{et al. 2002} \cite{905317} & Expectation-Maximization in HMM, Watch and Learn technique & N/A & Reduces the Problem of generalization by feature and the need for Testing and Training & Cannot recognize new emotions and gestures once learned. \\
\hline
Kapoor \textit{et al. 2005} \cite{inproceedings} & Mixture of Gaussian Processes and sensor fusion using two matrices of pressure sensors. & Custom dataset & 86\% accuracy in identifying interest in children given its nature to extract data from physical movement and not camera imaging & Requires a posture sensing chair with placed cameras. Tested only on children of younger age group. \\
\hline
Mitra \textit{et al. 2007} \cite{4154947} & HMM, Condensation Algorithms, FSM, Connectionist Approach & N/A & Multiple approaches for gesture recognition presented, along with parameters used in the implementation. & Focuses primarily on Hand Gestures and Facial Recognition, inadequate for understanding emotional state in educational context.\\
\hline
Rautaray \textit{et al. 2012} \cite{Rautaray2015} & OpenCV, MATLAB, A Forge. NET, iGesture & N/A & Detailed comparison between all algorithms and hardware and software developed for vision-based gesture recognition & Focuses primarily on Hand Gestures, inadequate for understanding emotional state in educational context. \\
\hline
Grafsgaard \textit{et al. 2013} \cite{10.1007/978-3-642-39112-5_1} & Kinect Camera and Sensor & Custom dataset with extensive footage collection using Kinect & Investigates interdependencies among tutorial dialogue, posture, and gesture. Automated techniques to test reactions & Returns negative feedback when two hands are on the face, without context for how the hands are placed.\\
\hline
Hariharan \textit{et al. 2014} \cite{6821668} & Kinect 3D and IR sensor camera with OpenNI for detection & Not Required & Able to detect the gesture and posture of a student using Kinect. & Recognizes the raised hand of students, cannot determine their emotional state in a classroom environment. \\
\hline
Zaletelj \textit{et al. 2017} \cite{Zaletelj} & Kinect Sensor, Matlab, Various Classifiers & Custom datasets (A, B, C) - 7 features identified. 1560 samples A/B, 5670 samples C. & Specifically for attention and behaviour of students in class, upper body and head posture, also includes facial features & Developed with intention of use in offline education. Requires Kinect One sensor. \\
\hline
Klein \textit{et al. 2017} \cite{Klein} & CNN based on AlexNet & Wits Intelligent Teaching System Dataset & Dataset created for associated engagement levels to various gestures. 89.7\% overall accuracy after 20,000 training epochs. & High level of training required. Developed with the intention of use in offline education. \\
\hline
Naik \textit{et al. 2018} \cite{8821186} & Viola-Jones Face Detector, CNN and RNN,  & N/A & Proposes methodology to recognize emotions such as decision making, assurance, fear, shame, etc. even with hand covering face. & Provides a detailed study, but it is not backed by reasoning for why the proposed model is better. \\
\hline
Santhoshk-umara \textit{et al. 2019} \cite{SANTHOSHKUMAR2019158} & Feedforward Deep CNN (FDCNN) & Emotion dataset (University of York) and GEMEP Corpus & The model performs better compared to other baseline models with an accuracy of 95.4\% & The prediction is not well balanced, it predicts some emotions better than others. \\
\hline
Ashwin \textit{et al. 2019} \cite{ashwinIEEE} & Scale-Invariant Context-Assisted Single Shot CNN & Custom dataset of 4423 images & Considers multi-person analysis of engagement levels in classroom & Tested for an offline environment with Camera. Does not consider both behavioural and emotional engagement. \\
\hline
Nguyen \textit{et al. 2019} \cite{Nguyen2019} & CNN and transfer learning with YOLOv3 detection & Custom dataset of 1102 images in 8 classes & Model has an 88\% accuracy in identifying engagement through gesture and posture. & Used footage from camera set up in offline classroom, may not yield same accuracy in online environment.\\
\hline
Keshari \textit{et al. 2019} \cite{Keshari2019} & Feature-Level Fushion, MultiSVM & Amrita Emotion Database-2 (AED-2) with 100 images in 7 basic emotions & Elaborates on feature-level fusion of features in images. 94\% accuracy in Bi-modal architecture. & Dataset used is very small. \\
\hline
Behera \textit{et al. 2020} \cite{Behera2020} & DNN, LinearSVM, IntraFace, RMSProp Optimizer & Custom dataset by capturing head and shoulder of subject & Studies the implications of hand-over-face gestures on difficulty of subjects. Uses webcam for dataset creation. & Only male subjects in dataset. Relies on the automatic detection of faces using IntraFace for detecting HoF gestures. \\
\hline
Noroozi \textit{et al. 2021} \cite{Noroozi2021} & SVM, DNN, RCNN, DeepPose, PoseTrack & RGB, GEMEP-FERA, LIRIS-ACCEDE, UFCKinect & Analysis of methodology gesture recognition in emotion analysis. Highlights the lack of deep representations for body posture and output space of posture analysis. & Does not cover any specific applications or use cases related to affect recognition in education. \\
\hline
\end{tabular}
\end{table}

\subsection{EYE TRACKING}
The predominant models and approaches for eye tracking recognition are SVM, CNN, Random Forest (RF) and Naive Bayes. \cite{KLAIB2021114037} SVM is used in the medical field to detect the eye movement of dementia patients or dyslexic children. Given its proficient use in this industry, it becomes a good candidate for classroom recognition. \cite{8803590} CNNs are a vital tool for eye tracking in AR and VR when the speed of eye movement is rapid and the environment is immersive. \cite{Zemblys2018} Random Forest, a type of supervised learning algorithm based on decision tree algorithms. The applications lie in event detection, pupil localization and blinking detection. \cite{9558860} The Naive Bayes model classifies individual eye tracking patterns using basic eye tracking metrics. It has aided in the understanding of children's visual cognitive behaviour and in recognising autism spectrum disorder. Of these, the support vector machine is the most widely used, due to its ability of binary classification. Table \ref{lit-eye} consists of the work done previously for the detection of eye tracking and its use in emotion detection.

\begin{table}
\caption{\textbf{Related Works for Eye Tracking}}
\label{lit-eye}
\tiny
\setlength\tabcolsep{1.5pt}
\begin{tabular}{ |p{1cm}|p{1.25cm}|p{1.25cm}|p{2.25cm}|p{2.25cm}| }
\hline
\textbf{Author} & \textbf{Tools/Models} & \textbf{Dataset} & \textbf{Merits} & \textbf{Demerits} \\
\hline
L.B \textit{et al. 2016} \cite{KRITHIKALB2016767} & Viola-Jones, LPB & N/A & Provides an unbiased feedback to content creator based on eye-movement and head rotation, without the need for a database. &  Determines attention levels by visibility of eyes. Does not account for instances where the eyes are visible but not focused on the screen. \\
\hline
Cheng \textit{et al. 2017} \cite{cheng2017} & NeuroSky brainwave (MindWave Mobile) and webcam & N/A & Develops a real-time attention detection and feedback generation system for MOOC Courses. & Does not provide details on how the visual sensor works or how it tracks learners' attention. \\
\hline
Zemblys \textit{et al. 2018} \cite{Zemblys2018} & Random forest classifier, Scikit-learn library & Custom baseline dataset recorded using PlayStation Eye Camera & Model to predict if eye gaze is fixated, saccade, or other oculomotor event.  & Unbalanced dataset with 89\% of the samples tagged as fixations. Model requires high computation.  \\
\hline
Mu \textit{et al. 2019}\cite{10.1007/978-981-13-6681-9_14} & Tobii Pro X3-120 & N/A & Discusses visualization of learning path in online courses by employing eye-tracking. & Requires an external eye-tracking device for the system. \\
\hline
Kaakinen \textit{et al. 2020} \cite{Kaakinen2021} & N/A & N/A & Study on the eye-tracking as indicators of different aspects of education. & Does not offer a comparison between the different models mentioned and discussed. \\
\hline
Sharma \textit{et al. 2020} \cite{Sharma2020} & Heat maps, SMI RED 250 eye-trackers, NRMSE & Custom Dataset with 40 students & Detailed implementation on the attentiveness of students for MOOC videos. Implements AOI backtracks for higher accuracy of attention levels. & Uses performance on a single test as basis for determining model performance. Requires additional eye tracker. \\
\hline
Klaib \textit{et al. 2021} \cite{KLAIB2021114037} & Lists down multiple tools and models like CNN, SVM, etc. & N/A & Provides highly detailed information about the various tools for eye tracking & Has methods that are invasive or require clinical observation, both unfeasible for online environment. \\
\hline
Jarodzka \textit{et al. 2021} \cite{Jarodzka2021} & Model on teacher's cognition & N/A & Explains the various eye tracking fields in online education & Provides an explanation only of theoretical concepts\\
\hline
Wang \textit{et al. 2021} \cite{wang2021} & Reference to build a binary classifier (Bayes nets, SVM, and decision trees) for eye gaze data & N/A & Exhaustive review of various eye-tracking techniques and datasets & Needs a way to manage and analyze larger amounts of data  \\
\hline

\end{tabular}
\end{table}

\subsection{FACIAL RECOGNITION}
Emotion Recognition approaches span two major models that are infamous in most of the cues for affect recognition. Firstly the CNN model \cite{Shen2022} is extensively used when multiple emotions need to be recognized with very small differences in sample data. Another popular model is SVM. Here it performs binary classification of given data. \cite{10.1007/978-981-15-0058-9_45}. However multiple binary classifications can be performed to include the needs of multiclass classification requirements. Table \ref{lit-face} is an overview of the work done in it thus far.

\begin{table}
\caption{\textbf{Related Works for Facial Recognition}}
\label{lit-face}
\tiny
\setlength\tabcolsep{1.5pt}
\begin{tabular}{ |p{1cm}|p{1.25cm}|p{1.25cm}|p{2.25cm}|p{2.25cm}| }
\hline
\textbf{Author} & \textbf{Tools/Models} & \textbf{Dataset} & \textbf{Merits} & \textbf{Demerits} \\
\hline
Ashwin \textit{et al. 2016} \cite{7395609} & Haar cascades, SVM & Yale Face Database (YFD) Face Detection Dataset and Benchmark (FDDB) & High accuracy and no false positives & Requires discussion of performance measures for eLearning \\
\hline
Abeer \textit{et al. 2017} \cite{YANG20182} & Haar Cascades, Sobel edge, Neural Network classifier training & JAFFE database & Higher accuracy using haar compared to sorbel & Requires exploration of illumination and pose of image \\
\hline
Li \textit{et al. 2018} \cite{8577590} & OpenCV, Dlib, CNN & FER2013 & Uses blink frequency along with emotions & Sample size and accuracy not mentioned \\
\hline
Ma \textit{et al. 2018} \cite{8468741} & CNN & FER2013 & After developing the model tested practically and obtained feedback & Dataset used employs webcam, can be subjected to poor image set \\
\hline
El Hammoumi \textit{et al. 2018} \cite{8525872} & Face detection using Haar Cascades in OpenCV, CNN & CK+ and the KDEF databases and JAFFE Database & High accuracy around 97\% & Investigate emotion state on e-learning system \\
\hline
Montazer \textit{et al. 2019} \cite{IMANI2019102423} & Kernel Density Estimation (KDE) & FACS mentioned & Highly detailed survey on speech and visual AR & Creation of real databases lacking \\
\hline
Alkabbany \textit{et al. 2019} \cite{Alkabbany2019} & CNN for extracting FACS, Gabor Filters and SVM for eye movement feature vector & Custom dataset along with BP4D, CK+ for training. & Provides real-time feedback for education based primarily on facial features and aspects of eye-tracking. & Dataset indicates varying levels of engagement but does not offer any insight into the specific emotions being experienced. \\
\hline
Ramos \textit{et al. 2019} \cite{ramos2020classifying} & Gabor Filter, SVM & Philippines-based corpus & Detailed analysis on said model & Relatively low accuracy  \\
\hline
Zhou \textit{et al. 2020} \cite{zhou2020deep} & Russel's model,VGSS network, caffe & CK+, JAFFE, MUG, ISED, RaFD,OULU,AffectNet, CMU-PIE & Very varied dataset, innovative way of grouping & 4 quadrant system insufficient to determine diverse human emotions \\
\hline
Min \textit{et al. 2021} \cite{9527452} & CNN, SVM and HOG & Dataset with pertinent academic emotions &Shows research on datasets and models & dataset relatively smaller \\
\hline
Altuwairqi \textit{et al. 2021} \cite{altuwairqi2021student} & mini-Xception model, CNN, SVM & FER2013 & Explains a very good technique for emotion recognition pertaining to the education domain & Requires a comparsion with other existing models to analyze efficiency and performance \\
\hline
Atsawaraungsuk \textit{et al. 2021} \cite{atsawaraungsuk2022fast} & Curvelet Transform (CT) and Local Curvelet Transform (LCT) & N.A & Compares models in depth & Can explore more feature extraction models and expand to a larger dataset \\
\hline
Lee \textit{et al. 2021} \cite{lee2021unpacking} & Ekman’s Facial Action Coding System & Custom dataset & Complete model and analysis provided & Number of subjects and the amount of data were rather small \\
\hline
Shen \textit{et al. 2021} \cite{shen2022assessing} & SE-CNN & JAFFE, ck+, and RAF-DB & Detailed model with 500 epochs & Not for real-time analysis, uses static data  \\
\hline

\end{tabular}
\end{table}

\subsection{SPEECH AND VERBAL RECOGNITION}
Speech Recognition can be guided by various models that generate an Automatic Speaker Recognition system (ASR) or for raw speech recognition. To identify the emotions of a speaker however, first requires the knowledge of what emotions can be heard and perceived as. The most employed technique for ASR is the Hidden Markov Model \cite{Mustafa2018}. But popularity doesn't always imply highest performance in terms of output. Support Vector Machine (SVM) has also been used based on its effective discriminative classifiers for easier identification \cite{HOSSAIN201969}. Using the Gaussian Mixture Model (HMM) can create a unique identification for each person in the speech recognizer \cite{7975169}. The Convolution Neural Network can be used to build a predictive model based on a predetermined dataset for learning \cite{HOSSAIN201969}. Multi Layer Perceptron (MLP) Classifier is a type of neural network which has been refined over time using backpropagation technique and optimizations in the activation functions. MLP Classifier is one of the best suited model for Speech Emotion Recognition (SER) using Mel-Frequency Cepstrum coefficients (MFCC) features \cite{10.1007/978-981-15-9647-6_39}. Table \ref{lit-ver} shows the information gathered by studying previous works of speech based classification.

\begin{table}
\caption{\textbf{Related Works for Verbal Recognition}}
\label{lit-ver}
\tiny
\setlength\tabcolsep{1.5pt}
\begin{tabular}{ |p{1cm}|p{1.25cm}|p{1.25cm}|p{2.25cm}|p{2.25cm}| }
\hline
\textbf{Author} & \textbf{Tools/Models} & \textbf{Dataset} & \textbf{Merits} & \textbf{Demerits} \\
\hline
Zhang \textit{et al. 2007} \cite{4376833} & One-Class-in-One Neural network (OCON) & Phoneme-balanced Chinese words & Gives a detailed analysis for designing the model & Does not cover unconsious expressions \\
\hline
Chen \textit{et al. 2007} \cite{10.1007/978-3-540-72588-6_91} & Neural network with speaker independence & N/A & Diagrammatically represents the process of speech recognition in E-learning & recognition rate only 50\% \\
\hline
Bahreini \textit{et al. 2015} \cite{bahreini2016towards} & FILTWAM framework, SMO of WEKA & N/A & Focuses on voice recognition from the perspective of e-learning & Sample size of only 12 members\\
\hline
Cen \textit{et al. 2016} \cite{CEN201627} & SVM (learning algorithm), LPCC, PLP, MFCC & Custom dataset used & Can recognise real-time, as well as pre-existing data, confusion matrix yields mean efficiency of 78\% for real time data & Results may be affected by discontinuous verbal interaction with students and background noises \\
\hline
Basu \textit{et al. 2017} \cite{7975169} & SVM, GMM, MLP, RNN , KNN, HMM & A table with known datasets has been mentioned & Gives a breakdown of various steps and procedures involved in emotional classification of speech. & Performance of feature extraction not mentioned \\
\hline
Mustafa \textit{et al. 2018} \cite{mustafa2018speech} & HMM, SVM, GMM and NN & N/A & Detailed comparisons for all parameters required in SER & Should attempt cross lingual SER \\
\hline
Hossain \textit{et al. 2019} \cite{HOSSAIN201969} & SVM, ELM (fusion), CNN (audio and video signals), LBP and IDP & Big Data and eNTERFACE database & Very clear and detailed explanation of the models proposed and procedure followed. Were able to achieve high accuracies using fusion techniques. & Can extend to training on dataset that have background noise comparable with real-world environment \\
\hline
Tao \textit{et al. 2019} \cite{tao_huang_li_lian_niu_2019}& Specific ladder network, MLP, VAE, SVM & IEMOCAP & Better accuracy as compared to SVM when testing & Beneficial to performance improvement of only two emotions \\
\hline
Zia Uddin \textit{et al. 2020} \cite{UDDIN2020103775}& Neural Structured Learning (NSL) model based on Neural Graph Learning (NGL) & N/A & Compares mean recognition rates of ANN, CNN, LSTM, NSL models & Trained model does not include background noise as in real-life environment \\
\hline
Kaur \textit{et al. 2021} \cite{10.1007/978-981-15-9647-6_39} & CNN, KNN, random forest, MLPClassifier & Berlin database of emotional speech - german language & Comparison among many different models to arrive at the result & Uneven distribution of emotions in the database - 23\% angry (total 7 emotions) \\
\hline
\end{tabular}
\end{table}

\section{Datasets Available}
The open source and custom-made datasets available to be used for all cues have been enlisted. Along with the description of sample size and data constituencies. Moving forward in the paper the datasets used and addressed confirm to this table. The following Table \ref{datasets} gives the summary of datasets that are publicly available for emotion recognition using NV cues.

\begin{table}
\caption{\textbf{Summary of Datasets available}}
\label{datasets}
\setlength\tabcolsep{1.5pt}
\tiny
\begin{tabular}{ |p{2cm}|p{2cm}|p{4.25cm}| }
\hline
\textbf{Parameter} & \textbf{Dataset~} & \textbf{Sample size}\\
\hline
Gesture Recognition and Posture Recognition & Custom Dataset~ & Around 4000 images with 3 categories\\
\hline
Eye tracking Recognition & LPW & 66 videos consisting of 130,856 separate frames\\
 & GazeCapture & 1474 subjects and 2.5 Million frames\\
 & NVGaze & Synthetic data with 2 Million images, Manual with 2.5 Million\\
\hline
Facial Recognition & FER2013 & 35,887 grayscale images\\
 & JAFFE & 213 images of 10 japanese women\\
 & CK+ Dataset & 593 images from 123 subjects\\
\hline
Speech or Verbal Recognition & IEMOCAP & 2400 training samples\\
 & RAVDESS & Speech Audio-Only Files (16bit, 48kHz.wav), 7356 files\\
\hline
\end{tabular}
\end{table}

For Gesture and Posture recognition, due to the lack of publicly available datasets, a custom dataset was created with around 4000 images in 3 categories identified as slouching, writing and sitting upright. Eye tracking has multiple data sets which could be used for tracking where the pupil seems to be directed, however none of these provide a classification as to what the different positions of pupils mean. Some of the datasets available are Labelled Pupils in the Wild (LPW) \cite{Tonsen_2016} for the implementation in pupil detection algorithms. GazeCapture \cite{gazecapture} a dataset created by MIT for the purpose for building a software that could be used via remote electronic devices like phones and tablets for eye tracking. NVGaze, created by NVIDIA \cite{nvgaze} for near-eye gaze estimation in infrared lighting. The FER2013 dataset \cite{fer2013}, created by Stanford provides a wide array of images classified into the 7 most commonly identified emotions for Emotion Recognition - happy, sad, angry, afraid, surprise, disgust, and neutral. It is an extremely comprehensive dataset for the purpose of Facial Expressions Recognition, however the accuracy of identifying emotions is comparatively low. The Japanese Female Facial Expressions (JAFFE) \cite{jaffe} dataset only consists of data based on Japanese women and hence could not be accurately used in this setup. Another dataset, the Extended Cohn-Kanade (CK+) dataset \cite{ckplus}, determines the basic emotions based on various action units identified in the facial expressions. It accounts for both posed and non-posed expressions. For Verbal and Speech Recognition in Affect Recognition, one of the commonly identified datasets is the Interactive Emotional Dyadic Motion Capture or the IEMOCAP \cite{iemocap} dataset. It consists of approximately 12 hours of data, with 151 videos and recorded audio. It can detect 9 emotions, namely angry, excited, fear, sad, surprised, frustrated, happy, disappointed and neutral. The Ryerson Audio-Visual Database of Emotional Speech and Song (RAVDESS) \cite{ravdess} dataset consists of 7356 only audio files for speech recognition. It consists of both speech and song audio clips, depicting different emotions.

\section{Implementation}
This section includes the implementation details of the models identified for each cue from the literature survey. The dataset used for different cues is also highlighted in this section. Based on the metrics such as accuracy, the best suited model is selected for proposing the multimodal architecture in the next section.

\subsection{GESTURE AND POSTURE RECOGNITION}
From the extensive literature survey the most prominent models identified are listed in Table \ref{mod-1}. The models along with the merits and demerits are considered and as identified from the literature survey, most researchers have suggested SVM and CNN models for emotion recognition based on gesture and posture. In this section, the performance of the two models on our custom dataset is compared. 

\begin{table}
\caption{\textbf{Models Identified for Gesture and Posture}}
\label{mod-1}
\setlength\tabcolsep{1.5pt}
\tiny
\begin{tabular}{ |p{1.5cm}|p{4cm}|p{3cm}| }
\hline
\textbf{Model} & \textbf{Merits} & \textbf{Demerits} \\
\hline
Hidden Markov Model (HMM) & Ability to recognize spatial and temporal behaviours.
Unsupervised learning model hence it does not require heavily annotated data.
 & Assumes a transition that depends only on the current and previous state. 
Unable to recognize unseen test data correctly in the evaluation phase.\\ 
\hline
Support Vector Machine (SVM) & Used in the case of a clear demarcation between the classes. 
More effective when it comes to multi dimensional spaces, especially cases where the number of dimensions is greater than the number of samples.
Relatively memory efficient.
 & Provides good accuracy only on small training data hence is not sensitive.\\ 
\hline
Kinect Sensor & Fast and accurate optical sensor for extracting 3D Information compared with other optical depth sensors
 & Algorithms for each gesture need to be in depth.
Requires really good camera footage.
\\ 
\hline
Convolutional Neural Network (CNN) & Lesser pre-processing, lower training time, and a smaller dataset
Automatically detects the important features without any human supervision
 & Large training data needed.
Doesn’t encode the position and orientation of objects.
\\ 
\hline
\end{tabular}
\end{table}

\subsubsection{Dataset Description} 
{The implementation of gesture and posture cue utilised the custom dataset that is made by identifying three main postures - Sitting upright, Slouching and Writing. The dataset consists of 4000 images that are of 10 subjects in variable environments so as to increase variations in training, as well as short clips taken from Youtube videos.}

\subsubsection{Convolutional Neural Network (CNN)}
\begin{figure}
\centering
\includegraphics[width=\linewidth]{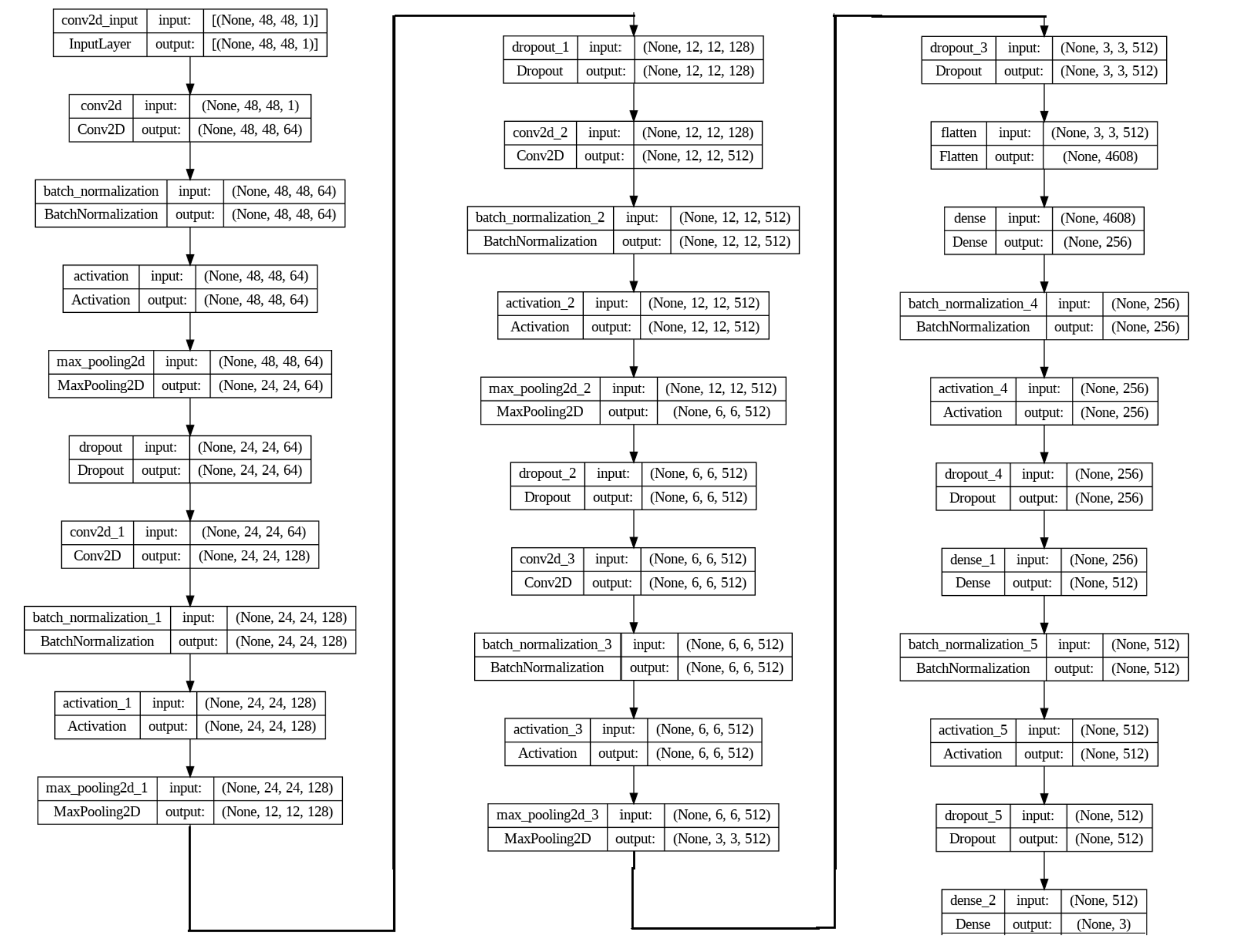}
\caption{\textbf{CNN Layer Architecture for Gesture and Posture}}
\label{fig:cnn-gesture}
\end{figure}

{CNN is a deep neural network in deep learning. The implementation involved the following architecture and is illustrated in Figure \ref{fig:cnn-gesture}. The sequential architecture of the CNN model has the first four CNN layers with Convolution, Batch normalisation, Max pooling and Dropout. These are followed by a Flatten layer and two fully connected layers. 
The mathematics of the convolutional layer and the training process involves the concept of convolution, forward and backward propagation. Let $X$ of size $N\times N$ be the input to the convolution layer. In this layer the filter or kernel chosen is $f$ of size $M\times M$. The convolutional layer output will be of size $(N-M+1)(N-M+1)$. Let $i, j$ represent the row and column corresponding to the indexes of the result matrix, $y$ which is the feature map. The calculation for each cell of the feature map is done using the formula below:}

\begin{equation}
\begin{split}
y[i,j]=(x*f)[i,j]= \sum_{k}^{M-1}\sum_{l}^{M-1} f[k,l]x[i+k,j+l]\label{eq} 
\end{split}
\end{equation}

\begin{figure}
\centering
\includegraphics[width=\linewidth]{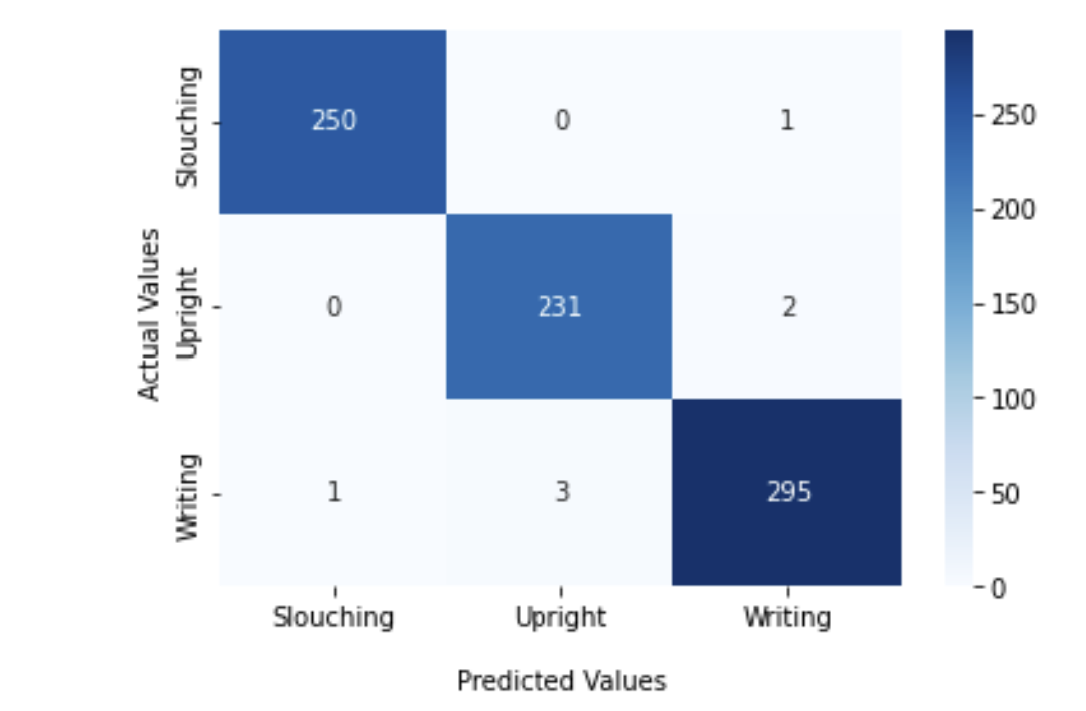}
\caption{\textbf{Confusion Matrix for Gesture and Posture Classification using CNN}}
\label{fig:cnn-gesture-2}
\end{figure}

\begin{figure}
\centering
\includegraphics[width=0.9\linewidth]{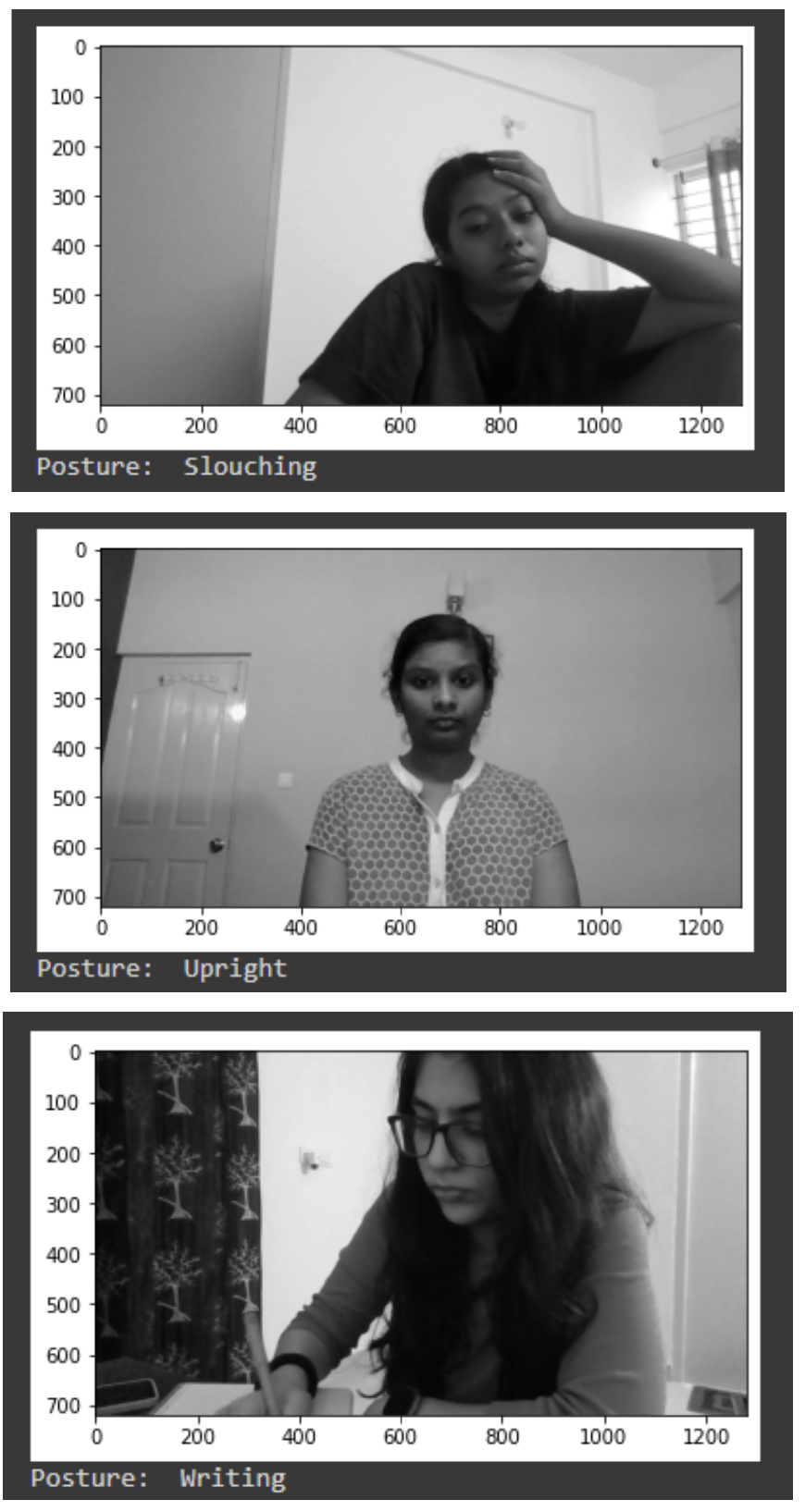}
\caption{\textbf{Gesture and Posture model testing: Slouching, Upright, Writing}}
\label{fig:cnn-gesture-1}
\end{figure}

Figure \ref{fig:cnn-gesture} demonstrates the CNN layer Architecture for Gesture and Posture. The CNN model gave a validation accuracy of 95.96\% on our custom dataset. These images were randomly picked and classified into 80-20\% testing and training. Figure \ref{fig:cnn-gesture-1} shows the classification on test sample images for different postures. Figure \ref{fig:cnn-gesture-2} shows the confusion matrix for CNN model. The confusion matrix of a model refers to a table created for comparing the predicted label of an image as opposed to the actual label. It consists of an $N\times N$ table where N refers to the total number of classes for classification, in this case, 3. The rows represent predicted class values while the column represents actual class values. As depicted in the matrix, the number of images which had the actual and predicted label as ‘slouching’ is 250, which is also the True Positive (TP) value for ‘slouching’. There is a single image labelled slouching that was falsely classified as ‘writing’, and refers to the False Negative (FN) value for ‘slouching’. More such inferences can be derived from the confusion matrix such as accuracy, precision, recall, f1 score, etc.

\subsubsection{ Support Vector Machine}
\begin{figure}
\centering
\includegraphics[width=\linewidth]{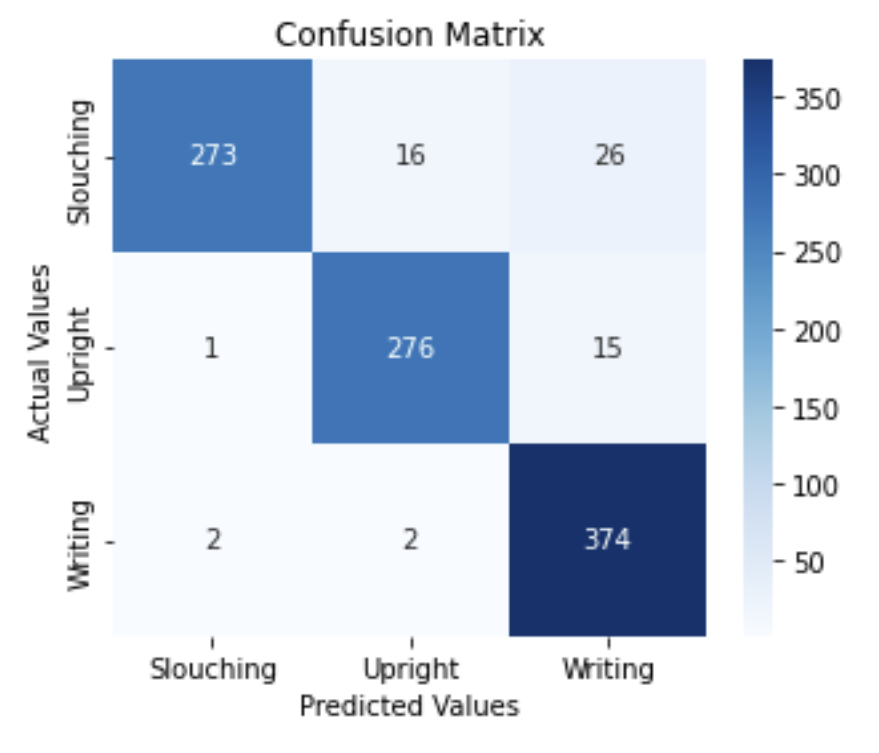}
\caption{\textbf{Confusion Matrix for Gesture and Posture Classification using SVM}}
\label{fig:cnn-conf-mat}
\end{figure}

{Supervised machine learning (SVM) has its applications in solving classifications and regression problems. SVM is used to perform a multiclass classification in order to perform emotion recognition.
For a training set 
\begin{equation}
\begin{split}
T = {(x_i, y_i)}\label{eq1} 
\end{split}
\end{equation}
 
where $x_i$ denotes the input vector and $y_i$ denotes the required output, the goal of a support vector machine is to find an optimal hyperplane such that 
\begin{equation}
\begin{split}
{W}^{t}x + b = 0 \label{eq2} 
\end{split}
\end{equation}

where $W$ is the adjustable weight vector, $b$ is the bias and $x$ is the input vector. For some values of $W$ and $b$, there is a margin of separation between the optimal hyperplane and the closest data point, denoted by $\rho$ . In a binary classification problem, the goal is to maximise the margin $\rho$ . However, 
\begin{equation}
\begin{split}
\rho=\frac{2}{||W_0||}\label{eq3} 
\end{split}
\end{equation} 
and hence, maximising the margin also implies minimising the weight vector W0. 
This implementation of SVM uses the RBF function in the sklearn library. For multi-class classification, Support Vector Classification (SVC) uses a "one-versus-one" or a “one-versus-rest” technique. The default implementation in sklearn is “one-versus-rest” where the hyperplane separates the class in consideration from all the others. This consists of n classifiers for n classes. In “one-versus-one” technique, there are a total of \begin{equation}
\frac{n(n-1)}{2}
\end{equation} classifiers built, each of which trains data from two classes. 
Radial Basis Function (RBF) is the kernel function used in this implementation. When using the RBF kernel to train an SVM the main parameters considered are C and gamma ($\gamma$). The parameter C is essentially used for balancing the prospect of wrong classification of the training samples due to the simplicity of the decision surface. If the C value is high it will correctly identify all the training instances, but a low C will smooth the decision surface. The parameter on the other hand, indicates the power of a unit training sample. The higher the value of $\gamma$, the samples must be equally close to be influenced. The RBF function is given by
\begin{equation}
\begin{split}
f(p,q) = exp(-{\gamma||p-q||}^{2})\label{eq4} 
\end{split}
\end{equation} 

Where $p$ and $q$ are two points in between which the distance score is calculated and hence converted to higher dimensional mapping, which makes computation easier.
The accuracy obtained was 93.7\% for linear SVM with the C parameter set to 1 and gamma set to the default setting. The implementation's confusion matrix is presented in figure \ref{fig:cnn-conf-mat}. 

While both the models gave a high accuracy for the dataset, owing to the higher accuracy achieved by the CNN model, it was chosen for the implementation of emotion recognition by gesture and posture.
}

\subsection{EYE TRACKING}
The Table \ref{mod-2} summarises the different models identified to be popular for eye tracking from the literature survey. Based on the merits and demerits of each model, SVM is best suited as the application requires a binary classification. This section discusses the SVM model for eye tracking, applied to gauge its use in a video conferencing mode. The eye tracking model aims to predict whether the student is paying attention to the screen or not based on pupil movement if they are looking away from the camera. The implementation is explained in further detail below.

\begin{table}
\caption{\textbf{ Models Identified for Eye Tracking}}
\label{mod-2}
\setlength\tabcolsep{1.5pt}
\tiny
\begin{tabular}{ |p{1.5cm}|p{3.5cm}|p{3.5cm}| }
\hline
\textbf{Model} & \textbf{Merits} & \textbf{Demerits} \\
\hline
Support Vector Machine (SVM) & Good response to stimulus and clear classification
 & Requires large, highly annotated dataset\\ 
\hline
Naive Bayes & Better for smaller sample sizes that require faster training.
 & Feature interactions cannot be incorporated as each of them is evaluated independently\\ 
\hline
Random Forest (RF) & The features can be used without scaling, centering, reducing or transforming them.
 & Cannot directly learn contextual information and sequences from the raw data. 
Also need post-processing before giving the final output.

\\ 
\hline
Convolutional Neural Network (CNN) & Provides accurate results despite a small training set.
 & Has dependable variables that control the accuracy of results like sample size, feature selection and data analysis.
\\ 
\hline
\end{tabular}
\end{table}

\subsubsection{Support Vector Machine (SVM)} 
\begin{figure}
\centering
\includegraphics[width=\linewidth]{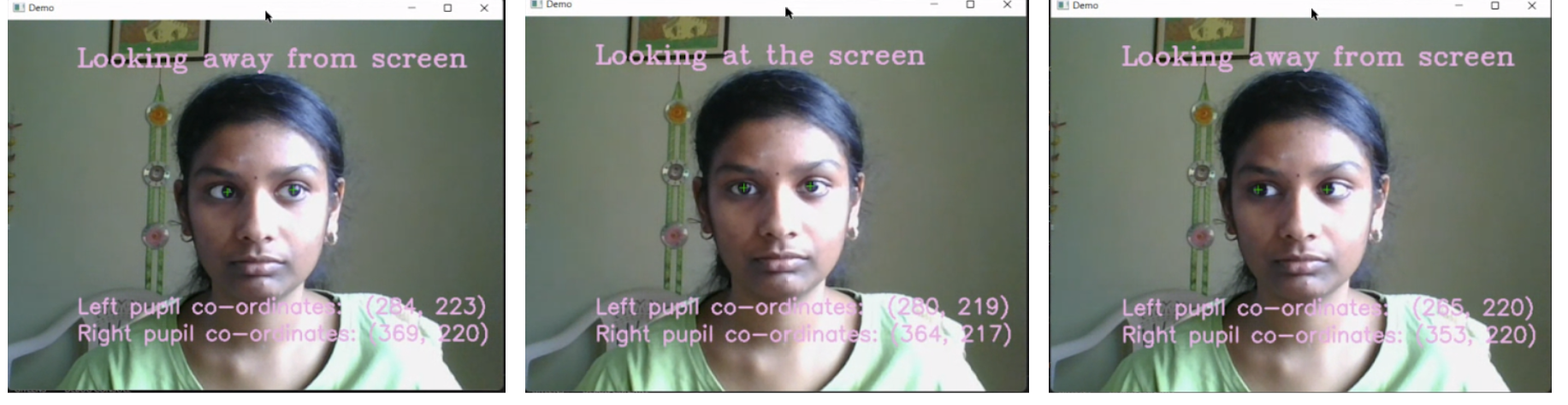}
\caption{\textbf{Eye Tracking positions: looking away from screen, looking at screen}}
\label{fig:eyetrack-img}
\end{figure}

{For eye-tracking, the OpenCV library is used for image processing and capturing real time data, along with the dlibs library for face recognition using the $frontal\_face\_detector$. The left and right coordinates of the pupils can be identified and based on the coordinate values, the gaze direction can be realised. This is classified into two categories: looking away from the screen and looking at the screen. The $frontal\_face\_detector$ works on the HOG (Histogram of Oriented Gradient) and Linear SVM that promises accuracy and high efficiency. Since SVM is primarily used for Binary Classification, the two categories are easily demarcated. Figure \ref{fig:eyetrack-img} shows the OpenCV frame snapshots of the two categories and respective pupil coordinates. The calculation is done based on the horizontal ratio calculated by the screen resolution size and absolute left and absolute right values taken as extremes.
}

\subsection{FACIAL RECOGNITION}
From the literature survey, the most preferred models are listed in Table \ref{mod-3} along with the merits and demerits of each model. Based on the characteristics of the specified models, the best-suited models for facial recognition are identified as Convolutional Neural Network (CNN) and Support Vector Machine (SVM). This section shows the respective implementation of models which classify emotion as either Bored, Confused, Frustrated, Interested or Neutral. 
\begin{table}
\caption{\textbf{ Models Identified for Facial Recognition}}
\label{mod-3}
\setlength\tabcolsep{1.5pt}
\tiny
\begin{tabular}{ |p{1.5cm}|p{3.5cm}|p{3.5cm}| }
\hline
\textbf{Model} & \textbf{Merits} & \textbf{Demerits} \\
\hline
Support Vector Machine (SVM) &Works comparatively better for well demarcated classification categories.
 & Unfit for larger datasets with multiple classifications with subtle differences\\ 
\hline
Convolution Neural Network (CNN) & Given the surplus data corpus available, can make a very sensitive model with high accuracy.
 & Multiple datasets with different formats need to be normalised and brought to same level.\\ 
\hline

\end{tabular}
\end{table}
\subsubsection{Dataset Description} 
{The dataset used is the FER2013 with over 35,000 greyscale images of various facial expressions. FER2013 dataset, created by Stanford provides a wide array of images classified into the 7 most commonly identified emotions for Emotion Recognition - happy, sad, angry, afraid, surprise, disgust, and neutral. For the purpose of the classroom environment, the available 7 categories are classified into 5 emotions - bored, confused, frustrated, neutral, and interested. The mapping is done by labelling happy and surprised as interested category, sad as bored, angry and disgust as frustrated, afraid as confused and neutral as neutral category.}
\subsubsection{Convolutional Neural Network (CNN)}
{
\begin{figure}
\centering
\includegraphics[width=\linewidth]{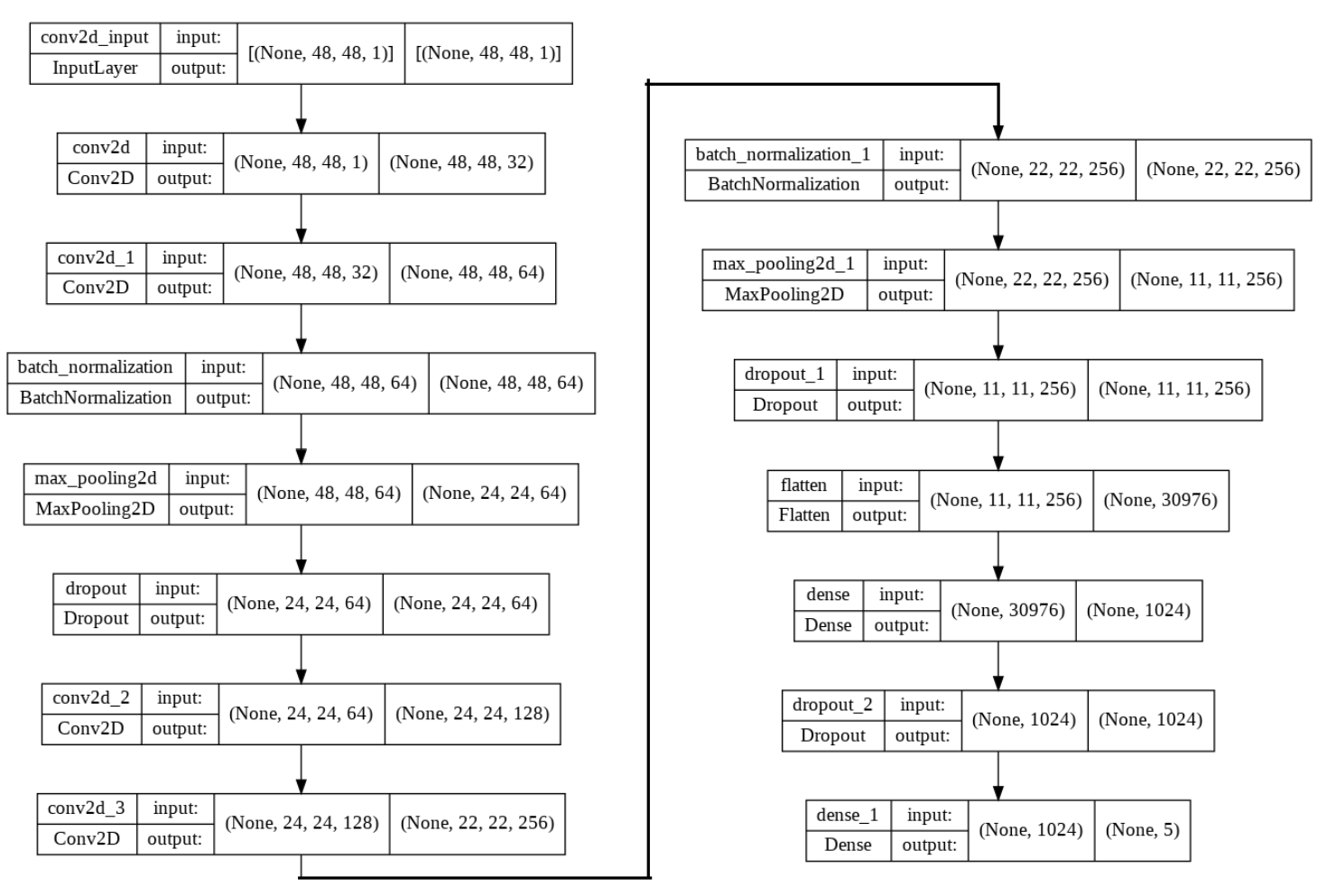}
\caption{\textbf{CNN Layer Architecture for Facial Recognition}}
\label{fig:facial-img-arch}
\end{figure}
The implementation involves the following architecture as illustrated in figure \ref{fig:facial-img-arch}. The sequential architecture of the CNN model has the 1st three CNN layers with Convolution, Batch normalisation, Max pooling and Dropout. These are followed by a Flatten layer, a fully connected layer, a dropout layer and finally a fully connected layer which classifies based on the 5 emotions. The training accuracy is 71.95\% and the validation accuracy is 65.07\% after running the model for 30 epochs. The difference between training and validation accuracy is simply due to overfitting because of the extremely large size of training data, i.e., over 35,000 images.
}
\subsubsection{Support Vector Machine (SVM)}
{
\begin{figure}
\centering
\includegraphics[width=\linewidth]{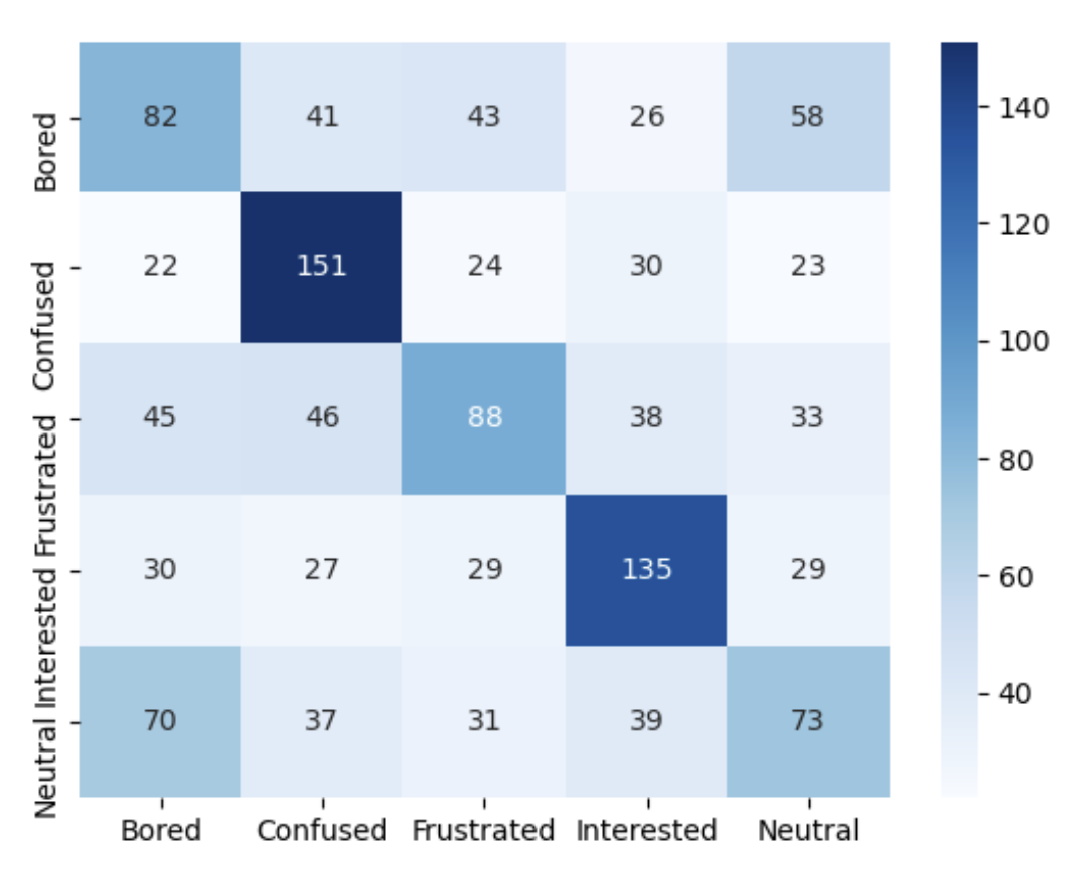}
\caption{\textbf{Confusion Matrix for Facial Recognition using SVM}}
\label{fig:facial-img-conf}
\end{figure}
The SVM model yields an accuracy of 42.32\% with the C parameter set to 1 and gamma set to the default setting for facial recognition. The implementation's confusion matrix is presented in figure \ref{fig:facial-img-conf}. Hence comparing the results of both implementations, the CNN model was selected for the emotional recognition of facial expressions.
}

\subsection{SPEECH AND VERBAL RECOGNITION}
Table \ref{mod-4} lists the models researchers used for speech recognition as identified from the literature survey. Based on the merits and demerits, the best suited model for speech recognition is identified to be CNN. In this section, the CNN model is built that classifies verbal emotions based on five categories namely, Bored, Confused, Frustrated, Interested or Neutral. 

All the cues have taken SVM as a baseline architecture to compare the performance of one another. Even when conducting the literature survey, it is brought to notice that Support Vector Machine aids in binary classification and can be extended to multiple classifications quite easily. The use of one baseline architecture shows how the model performs based on the available cue dataset, thereby indicating the strength of the dataset collected.

\begin{table}
\caption{\textbf{ Models Identified for Verbal Recognition}}
\label{mod-4}
\setlength\tabcolsep{1.5pt}
\tiny
\begin{tabular}{ |p{1.5cm}|p{3.5cm}|p{3.5cm}| }
\hline
\textbf{Model} & \textbf{Merits} & \textbf{Demerits} \\
\hline
Hidden Markov Model (HMM) &Most used technique for automated speech recognition (ASR)
 & Does not provide high performance ASR Systems.\\ 
\hline
Support Vector Machine (SVM) & Powerful discriminative classifier. 
Guaranteed to converge to the least value of associated cost function

 &Does not perform well with noise i.e, target classes are overlapping.\\ 
\hline
Gaussian Mixture Modelling (GMM) & Unique identity can be built for each person from extracted features.
 & Recognizes an individual from a spoken sentence. 
Not to detect emotions.
\\ 
\hline
Convolution Neural Network (CNN) & Can build predictive neural network models or hybrid HMM-CNN models.
 & Lack of availability of large dataset classifying emotion recognition speech.\\ 
\hline
Multi Layer Perceptron (MLP) Classifier & Can learn non-linear relationships.
Robust to noise and variance in speech input.

 & Works well only when careful feature engineering is employed, which can be time-consuming.\\ 
\hline

\end{tabular}
\end{table}

\subsubsection{Dataset Description} 
{The RAVDESS Emotion Recognition Dataset was adopted for the purpose. The speech audio-only files were used where 24 professional vocal actors voiced two north american sentences in various tones.}
\subsubsection{Convolutional Neural Network (CNN)}
{The librosa library in python for audio and music analysis was used. All features were extracted from the sound files in .wav format. The three features extracted were ZCR (Zero Crossing Rate), RMS and MFCC (Mel Frequency Cepstral Coefficients). The dataset was then split into testing and training sets to identify the accuracy of the model. The test accuracy was found to be 85\% and training accuracy is 95\%. On testing with some sample audio files it was observed that the model does not provide accurate classification thereby intimating that it is overfitting the trained dataset.}
\subsubsection{MLPClassifier}
\begin{figure}
\centering
\includegraphics[width=\linewidth]{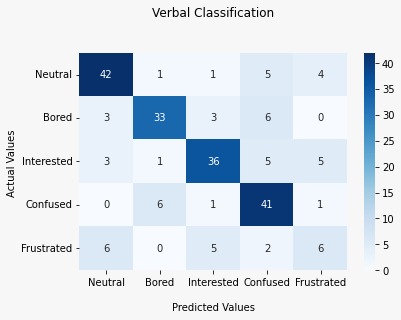}
\caption{\textbf{Confusion Matrix for Speech Recognition using MLPClassifier.}}
\label{fig:conf-mat-verbal}
\end{figure}

{MultiLayer Perceptron Classifier maps input dataset appropriately with the desired output using a feedforward artificial neural network model. It employs a nonlinear activation function and can have one or more nonlinear hidden layers.

The features extracted from the sound file are MFCC, Chroma and Mel. The parameters used for the classifier are:

\begin{equation}
\begin{split}
 model = &MLPClassifier(alpha=0.01, batch\_size=256, \\ &epsilon=1e-08, hidden\_layer\_sizes=(300,), \\&learning\_rate='adaptive', max\_iter=500)\label{eq5}
\end{split}
\end{equation}

The test accuracy obtained for the above trained model is 73.15\%. Then it was sampled against our own vocals and predictions were matched. Figure \ref{fig:conf-mat-verbal} is the confusion matrix for Speech Recognition using MLPClassifier. }

\section{Multimodal Proposed Architecture}\label{sec11}

Multimodal Affect Recognition refers to interpreting human emotions, where modalities refer to the various stimuli and data inputs that can be used to determine the final output. This section expands on the previous works in this field and their limitations. Further, it elaborates on an approach that uses decision-level fusion to classify student emotions in an online educational setting.

\subsection{RELATED WORKS}
The multimodal architecture’s effectiveness depends on the fusion strategy employed for combining different modalities. In general, there are three fusion strategies. Firstly the feature-level fusion strategy is where the feature vectors of different modalities are concatenated as input to the classifier \cite{zhao}. Feature-level fusion can be simple concatenation, CNN-based fusion, single autoencoder or multi autoencoder fusion. \cite{subramanian} The second approach is the model-level fusion approach in which the correlations between the modalities are identified to obtain feature vectors from the modules and this is the input to the classifier \cite{Castellano2008}. The last approach is the decision-level fusion or late fusion approach. \cite{patania} The classifiers are trained independently for each modality and the final emotion is predicted using the majority voting technique, best probability technique or weight multiplicative combination technique. \cite{kawde} proposes Bayesian inference classification-based decision-level fusion that uses joint probabilities for classification. An overview of the work done in this field is given in Table \ref{lit-mul}.

\begin{table}
\caption{\textbf{Related Works for Multimodal Affect Recognition}}
\tiny
\label{lit-mul}
\setlength\tabcolsep{1.5pt}
\begin{tabular}{ |p{1cm}|p{1.25cm}|p{1.25cm}|p{2.25cm}|p{2.25cm}| }
\hline
\textbf{Author} & \textbf{Tools/Models} & \textbf{Dataset} & \textbf{Merits} & \textbf{Demerits} \\
\hline
Castellano \textit{et al. 2008} \cite{Castellano2008}& Neural Networks & Custom Dataset (GEMEP corpus) & Clear comparison between unimodal and multimodal models with two fusion techniques & The performance of the unimodal classifiers was 48.3\%, 67.1\% and 57.1\% and each showed a disparity in recognition of particular emotion. Despair emotion is not recognised effectively even in fusion models.\\
\hline
Paleari \textit{et al. 2010} \cite{5518574}& Bayesian Classifier & eNTERFACE'05 database & Extensive study on three feature sets (i.e. coordinates, distances, and audio) for the task of real-time, person independent, emotion recognition & No justification given for selection of Neural Networks for training\\
\hline
Banda \textit{et al. 2011} \cite{10.1007/978-3-642-24571-8_21} & Support Vector Machine (SVM) & Extended MindReading audio DVD & Pairwise comparisons are done and based on the count the each pairwise win, the emotions are displayed & The formulated decision fusion creates a loss of information between video and audio modalities\\
\hline
Panning \textit{et al. 2012} \cite{6335662}& Linear Filter and PCA & Last Minute Corpus (LMC) & Identified that not all modalities contribute to classification & High variation in results that range from 40-90\% based on different subjects considered\\
\hline
Gonzalez-Sanchez \textit{et al. 2013} \cite{6681524}& Pleasure, Arousal and Dominance (PAD) Model & EEG Headset and webcam & First attempt to indicate how multimodal affect recognition can be adjusted for diverse range of projects & Study conducted is generic and results are not clearly defined\\
\hline
Osman \textit{et al. 2017} \cite{Falk17} & Lists the various cues and models that work with them respectively & List of publicly available datasets for Emotion Recognition & Highlights the different approaches to dealing with modalities & N/A\\
\hline
Kawde \textit{et al. 2017} \cite{8358500} & Stack auto-encoder (SAE) and Deep Belief Network (DBN) & DEAP Database & Clear detailed analysis with mathematical modelling and graphical representation & Only focuses on three emotional states (valence, arousal, and dominance) and does not consider other emotions\\
\hline
Zhao \textit{et al. 2018} \cite{8470385} & LSTM applied for temporal facial extraction using DenseNets, Residual Network (ResNet) for visual context and RF and MLP for emotion classification & CHEAVD 2.0 & It outperforms the MEC2017 baseline system by a margin of 19.06\% & Low accuracy of 45\% from testing the proposed model might provide inaccurate results\\
\hline
Dewan \textit{et al. 2019} \cite{Dewan2019} & Multiple models explored for each cue. & N/A & A Review on the various techniques that can be implemented for exploring Multimodal Affect Recognition in Education. & No comparative analysis of the various models suggested. \\
\hline
Ashwin \textit{et al. 2020} \cite{TS2020334} & CNN & Custom Dataset mixed with classroom environment and facial data & Comprehensive dataset. 76\% accuracy & Does not provide the feasibilty and cost of preparation and annotation of large dataset. Does not consider verbal cues\\
\hline
Chan \textit{et al. 2020} \cite{Chan2020} & HyperFace - head direction. DockerFace - detect faces, OpenPose - All poses & Custom Dataset - works mostly on recognizing features & Multiple classroom aspects such as noise level, teacher speaking volume, teacher-student interactions, etc. have been modelled & Does not take into account online learning\\
\hline
Halfon \textit{et al. 2020}\cite{doi:10.1080/10503307.2020.1839141}& OpenPose, a ResNeXt style architecture & AffectNet database. Semi-automatic prepared resource & Good Performance shown by SVR with RBF kernel (Anger), SVR with linear kernel (Sadness) and ELM (Anxiety, Pleasure). & Considers only two modalities - face and text\\
\hline
Cimtay \textit{et al. 2020} \cite{9195813}& CNN, InceptionResnetV2 & Facial - Cohn-Kanade+ face dataset, Radboud Faces Database, FacesDB and AffectNet, LUMED-2 and DEAP & Works on CNN model and decision trees- easy to understand approach & Cues considered hard to implement in educational environment as used EEG and GSR data modalities\\
\hline
Adiga\textit{ et al. 2020} \cite{9311566} & 1D and 2D CNN & FER-2013 dataset, RAVDESS & Clear explanation of multimodal approach with 2 method strategy & No justification for selecting Multimodal approach based on Decision tree and how it is effective\\
\hline

\end{tabular}
\end{table}

\begin{table}
\label{lit-mul-2}
\setlength\tabcolsep{1.5pt}
\tiny
\begin{tabular}{ |p{1cm}|p{1.25cm}|p{1.25cm}|p{2.25cm}|p{2.25cm}| }
\hline
\textbf{Author} & \textbf{Tools/Models} & \textbf{Dataset} & \textbf{Merits} & \textbf{Demerits} \\
\hline

Tzirakis \textit{et al. 2021} \cite{TZIRAKIS202146}& CNN, HRNet for Visual, LSTM for temporal dynamics & SEWA dataset & Uses text, visual and audio modalities. Good performance of proposed model, Code repository made open source to help researchers & Complexity of the proposed model limits its application capability\\
\hline
Sümer \textit{et al. 2021} \cite{9613750} & Attention-Net for head pose estimation and Affect-Net for facial expression recognition & TC3Sim game recordings & Considers both uni and bi-directional sequences & Feature Selection done without assigning weights to various cues\\
\hline
Ramakrish-nan \textit{et al. 2021} \cite{9353969}& CNN, Temporal Neural Network for integration, OpenFace, OpenPose & MET dataset, UVA dataset & Comparison of various models done over two datasets - TNN outperforms BiLSTM & Defined for offline education\\
\hline
Henderson \textit{et al. 2021} \cite{9597432} & BROMP, MTL, Cross-Stitch Networks, Feature Selection - Greedy Technique, FFNNs & TC3Sim game recordings & Considers both uni and bi-directional sequences & Feature Selection done without assigning weights to various cues\\
\hline
Bhattacha-rya \textit{et al. 2021} \cite{9397275}& SVM with a polynomial kernel & One-Minute-Gradual (OMG) Emotion Dataset & Accuracy measure listed in detail for each feature extracted & The modality effects differ across emotion types\\
\hline
Subramani-am \textit{et al. 2021} \cite{9445146} & SVM+ autoencoder & FER and Ck+, RAVDESS and SEED-IV & Four feature fusion approaches were attempted & EEG modality not applicable for online class environment\\
\hline
Patania \textit{et al. 2022} \cite{10.1007/978-3-031-06430-2_61} & CNN for feature extraction. RNN for affective behaviour modelling & RECOLA & Compares combination of various modalities & EEG hard to implement in online class environment\\
\hline

\end{tabular}
\end{table}

\subsection{PROPOSED ARCHITECTURE}
Upon performing the individual cue analysis, it can be verified that there exists no one model type that is best suited for all cues. Since all cues have different features that need to be considered, logically feature-level fusion may not be the appropriate approach. In comparison, given that the final emotion returned by each cue will be taken into account for combining the multiple modals, the best approach to follow would be the decision-level fusion. Decision fusion implies combining the results of several models into a single decision about the action taking place.

Currently, no such model exists that gives an output based on outputs received from each cue of the feature extraction. One technique that can be used is a decision tree. In a decision tree model, the result is based on defined criteria or parameters. Based on the criteria that matches, the next set of parameters is applied until the final decision is obtained. However, these decision trees are manually made and hence prone to human errors. It is also very costly to go through the entire process of making a decision tree for each picture and frame, therefore in a real-time environment it would be completely impractical. 

Another way of deciding the combined output to different cues could be to employ parallel decision-level fusion. Parallel implies that all the models run simultaneously and store their required outputs until all models are ready for consolidation. There is a high probability that there is a conflict between the emotions detected by different models. In such cases, various voting ensemble techniques can be employed. Majority voting, the majority vote of the individual models is taken as the final prediction. Another approach is weighted voting where individual models are assigned weights based on parameters such as the accuracy of the model.

Let us assume that $y_i$ is the output emotion of the $i^{th}$ student. For determining the decided emotion, a differential weight $w_j$ must be allotted to every $j^{th}$ cue considered in predicting $y_i$. Let $o_{i_1}, o_{i_2}, \dots o_{i_n}$ where n is the number of cues, denote the output for each modality considered for the $i^{th}$ output. Then the output $y_i$ may be calculated as
\begin{equation}
y_i = \sum_{j=1}^{n} w_jo_{i_j}
\end{equation}

But even though the best-identified approach is parallel decision-level fusion, most techniques within the branch require all parallel features to make decisions based on the same output labels. However, the output labels in our case are all different from each feature extraction. The decision however still can be made using a decision tree, but it becomes more costly. A proposal could be to use the matching decision-level fusion with some altercations. Each final emotion label can be given a matching vector. This matching vector is then compared with the output features to identify a soft vote of each of the matching vector items and then a score is provided. This score is calculated taking into account the weight allotted to various features and then a rank of all 5 emotions is made. The highest-ranking emotion label will be allotted as the final decision at the decision-level fusion.

\begin{figure}
\centering
\includegraphics[width=\linewidth]{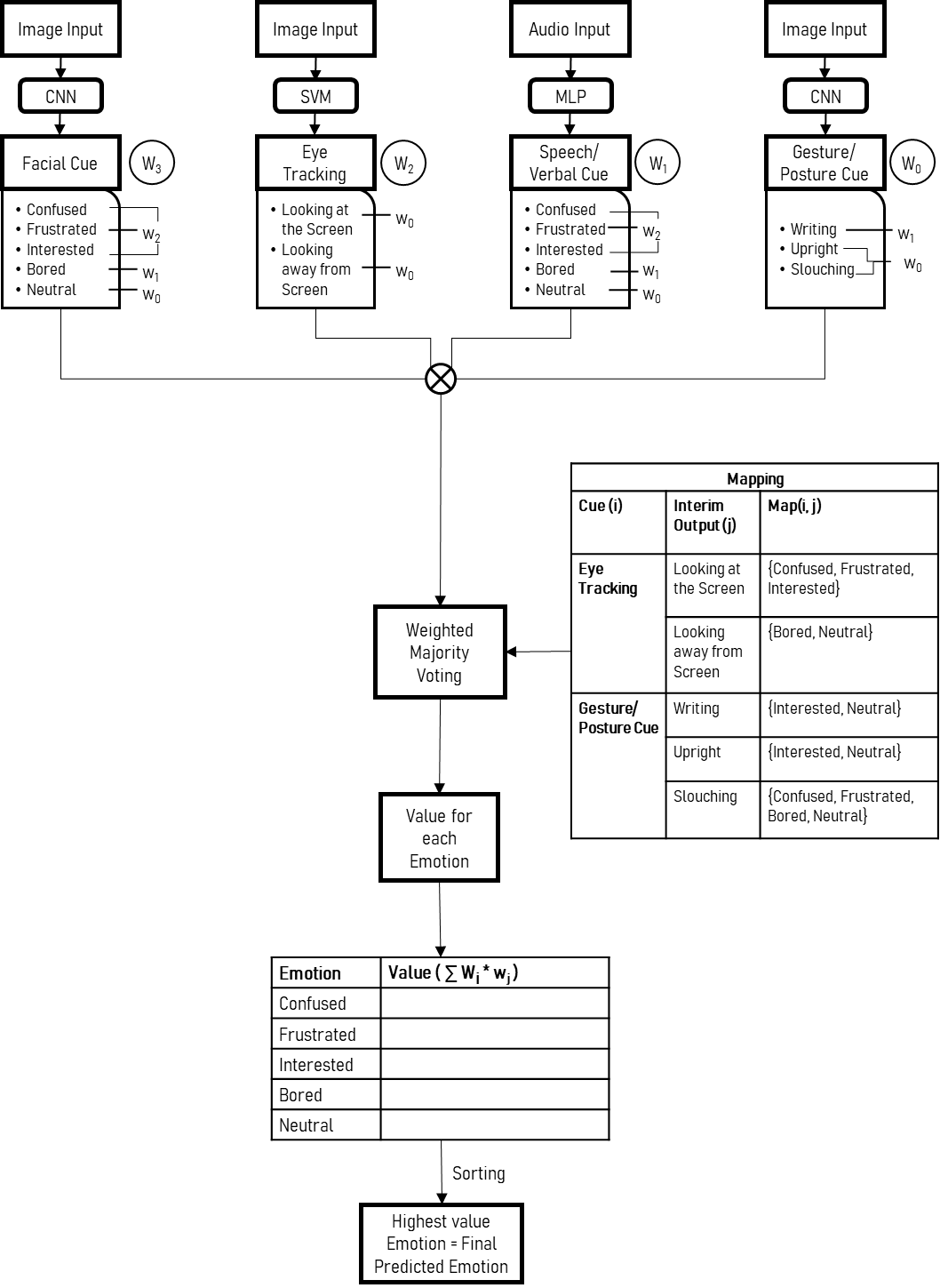}
\caption{\textbf{Multimodal Proposed Architecture}}
\label{fig:arch}
\end{figure}

Figure \ref{fig:arch} is a diagrammatic representation of the proposed architecture. The input is fed to each cue and the subsequent models have been selected based on the comparison conducted in the implementation in Section IV. The individual cues then provide the intermediate cue output response. Here each output is given a weight based on the priority considered for the cue from $W_0$ to $W_3$, the higher the weight value, the more the priority. The output from each of the cues is listed and subsequent sub-weights are allotted to each of the output values based on factors of the likelihood of accuracy and subjectivity. For the Eye-tracking cue and Gesture and Posture cue, the interim outputs are not reflective of their assigned emotion, hence it is compared to the assigned emotions to allot the same using the mapping function, $Map$. For the cue weights $W$ and their sub-weights $w$, the equation for calculated emotions is as follows, 

\begin{equation}
\begin{gathered}
\text { Final emotion }=\max _e\left(\sum W_i \times w_j\right), \\
\forall e \in \text { Emotions, } \\
\operatorname{Map}(i, j) \cap e \neq \varnothing
\end{gathered}
\end{equation}

Succeeding which a majority vote is conducted for each emotion, but with the calculated weights and the emotion with the highest value is deemed as the Final predicted emotion from the multimodal architecture.

% \begin{figure*}[h]
% \centering
% \includegraphics[width=\linewidth]{arch.png}
% \caption{\textbf{Multimodal Proposed Architecture}}
% \label{fig:arch}
% \end{figure*}

\begin{figure}
\centering
\includegraphics[width=\linewidth]{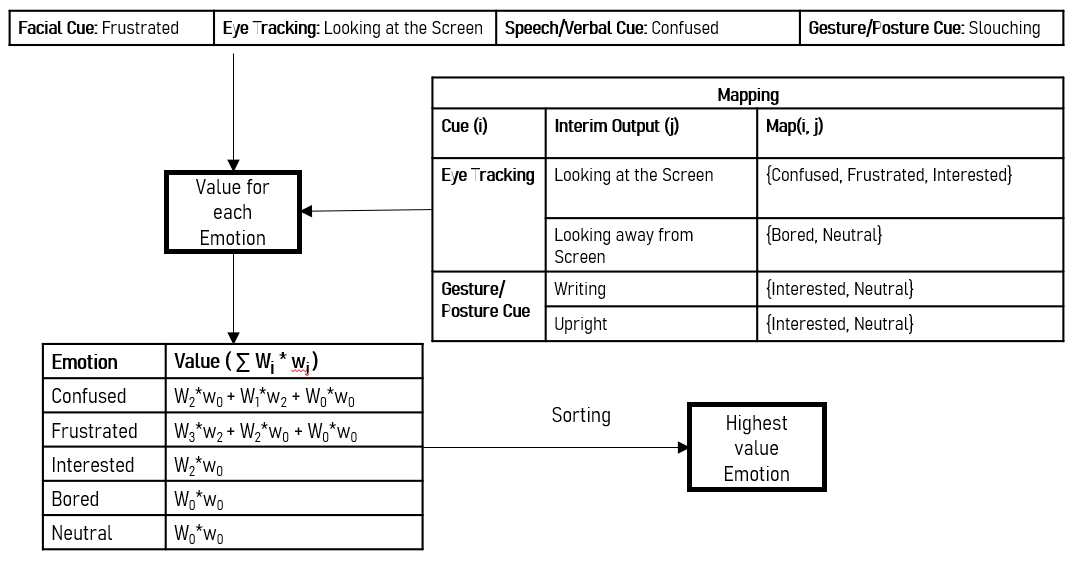}
\caption{\textbf{Example of Multimodal Emotion Recognition}}
\label{fig:arch-example}
\end{figure}

To better understand the proposed architecture, an example can be considered. Figure \ref{fig:arch-example} depicts a model example. The initial step indicates the predicted interim output from each model. For Eye Tracking, the emotions assigned to Looking at the Screen are Confused, Frustrated and Interested. So the required weights are multiplied by the cue weight and added to the respective emotions. 
\begin{equation*}
[Interested= W_2 w_0]
\end{equation*}

Similarly, the calculation is made for the Gesture and Posture cue and the value is added to the respective cue as shown in the table. Based on the weights, it becomes evident that for the vector $[Frustrated, Looking At Screen, Confused, Slouching]$ the output final predicted emotion is $[Frustrated]$.

% \begin{figure*}[h]
% \centering
% \includegraphics[width=\linewidth]{arch-example.png}
% \caption{\textbf{Example of Multimodal Emotion Recognition}}
% \label{fig:arch-example}
% \end{figure*}

\section{Conclusion}\label{sec13}

Summarising the outcomes of the cue-generated model analysis for detecting emotions in an online setting, it can be inferred that a multimodal approach that comprises a combination of multiple cues is the best approach for the accurate prediction of emotions. However, to get accurate comparisons with single cue models, it becomes imperative to extend the study by implementing the proposed model and testing it in a real-time environment to see how accurately assessments are made. The proposed model can also be applied to students with disabilities where cues can be better understood by affect recognition in a multimodal environment as it focuses on multiple aspects of emotion recognition. In this case, the weightage given to speech recognition can be reduced so as to have the other cues dominate. Another use case is found in providing quality education through such AI-driven remote proctoring tools, to students in remote areas to capitalise on the best aspects of urban education.

%%%%%%%%%%%% Supplementary Methods %%%%%%%%%%%%
%\footnotesize
%\section*{Methods}

%%%%%%%%%%%%% Acknowledgements %%%%%%%%%%%%%
%\footnotesize
%\section*{Acknowledgements}

%%%%%%%%%%%%%%   Bibliography   %%%%%%%%%%%%%%
%\normalsize
\bibliography{multimodal-preprint}

%%%%%%%%%%%%  Supplementary Figures  %%%%%%%%%%%%
%\clearpage

%%%%%%%%%%%%%%%%   End   %%%%%%%%%%%%%%%%
%\end{multicols}  % Method B for two-column formatting (doesn't play well with line numbers), comment out if using method A
\end{document}